\documentclass[10pt,letterpaper]{article}

\usepackage[utf8]{inputenc}
\usepackage[T1]{fontenc}
\usepackage{times}
\usepackage{graphicx}
\usepackage{booktabs}
\usepackage{amsfonts}
\usepackage{amsmath}
\usepackage{amssymb}
\usepackage{nicefrac}
\usepackage[nopatch=footnote]{microtype}
\usepackage{xcolor}
\usepackage{longtable}
\usepackage{multirow}
\usepackage{siunitx}
\usepackage{bm}
\usepackage{makecell}
\usepackage{subcaption}
\usepackage[hidelinks]{hyperref}
\usepackage{natbib}
\usepackage[capitalize,noabbrev]{cleveref}
\usepackage{placeins}
\usepackage{tabularx}
\usepackage{array}
\usepackage[bottom]{footmisc}
\usepackage{algorithm}
\usepackage{algorithmic}
\usepackage{geometry}
\title{Policy Network Diffusion}

\author{
Jiate Liu\thanks{Work done at Tencent in 2024.}
\and
Deheng Ye$^{*}$
}

\date{}

\begin{document}

\maketitle

\begin{abstract}

Diffusion models have shown remarkable progress in AI-generated content, including images and videos. In this paper, we explore whether diffusion models can generate the parameters of policy networks in reinforcement learning (RL), and whether the generated policies exhibit generalization capabilities. To achieve this, we propose Diffusion-based Conditional Policy Network Parameter Generation (DPPG), which consists of a Transformer-based autoencoder and a task-conditioned diffusion model. The autoencoder achieves a high compression rate by patching network parameters, while the task-conditioned diffusion model generates corresponding parameters conditioned on different environments. Our experiments demonstrate that with a single unified DPPG model, we can synthesize high-quality policies across a diverse set of tasks from different environment domains, including Classic Control, Atari games, and MuJoCo. Additionally, our method can generate networks with a larger number of parameters compared to previous approaches. Furthermore, we find that, in previously unseen environments, DPPG can generate parameters that achieve faster convergence and reach competitive performance, showcasing the potential future applications of parameter generation via diffusion.

\end{abstract}

\vspace{8pt}
\noindent\textbf{Preface.} When this work was originally conducted at Tencent in 2024, parameter generation via diffusion was largely viewed as a niche meta-learning curiosity. The field was dominated by supervised learning benchmarks, and the idea of synthesizing reinforcement learning policy networks directly from random noise received limited attention. However, the recent breakthrough that frontier models can directly edit the weights of open-source LLMs---effectively treating weight modification as a generative task---has triggered a sudden reevaluation of this paradigm. What was once dismissed as a toy problem is now recognized as a foundational capability: if we can generate or edit weights directly, we unlock a fundamentally new way to adapt, customize, and compose models without expensive from-scratch training. This paper stands as one of the first systematic explorations of diffusion-based parameter generation for decision-making policies, predating the current wave of interest by nearly two years. We hope that this revival serves as a reminder that seemingly premature ideas often arrive years before their time.

\section{Introduction}

Originating from non-equilibrium thermodynamics \citep{jarzynski1997equilibrium}, diffusion processes were introduced for denoising in unsupervised learning \citep{diffusion1} and soon refined to denoising diffusion probabilistic models (DDPM) \citep{ddpm, ddim}. Unlike traditional generative methods such as variational autoencoders \citep{vae, beta-vae} and adversarial generative networks \citep{gan}, diffusion models adopt a training paradigm consisting of a forward process and a reverse process. The superior generation quality of diffusion models has significantly advanced the progress of AI-generated content (AIGC), with unprecedented generation quality for images and videos \citep{stable-diffusion, dalle2, imagen-video}.

The successful applications of diffusion models in image and video synthesis inspire research on adapting diffusion models to generate other types of content. One such attempt is to synthesize model parameters using diffusion processes \citep{pdiff}. By directly fitting the optimized model parameter distribution, we can utilize diffusion models to sample network parameters for desired task requirements, known as a learning-to-learn paradigm \citep{meta-learning, maml}. Parameter generation using diffusion models typically shows promising results compared with prior approaches using hyper-networks \citep{hypernet} and variational autoencoders \citep{G.pt}. These works commonly focus on generating neural network parameters for ad hoc domains such as image classification \citep{pdiff, d2nwg}, tabular data prediction \citep{d2nwg, gpd}, and implicit neural fields \citep{hyperdiffusion}.

Nevertheless, existing work mainly considers generating parameters for neural networks in limited supervised learning tasks. In this paper, we explore whether this paradigm can be extended to more complex reinforcement learning tasks. Supervised learning tasks often yield relatively similar network checkpoints with a stable training process, whereas reinforcement learning tasks present more challenging conditions with diverse network structures and parameter distributions. For instance, there may be multiple optimal policies that solve a task in different ways, leading to a more diverse distribution for diffusion models to fit. Moreover, the differences in observation spaces and action spaces across reinforcement learning tasks also require more flexible and adaptive parameter generation methods.

Based on these requirements, we propose Diffusion-based Conditional Policy Network Parameter Generation (DPPG), a fully Transformer-based model that generates policy parameters for multiple reinforcement learning tasks. DPPG consists of an autoencoder and a latent diffusion model. The autoencoder is designed to compress policy parameters into a latent representation; we find that a high compression rate can be achieved by patchifying the parameters and feeding them into a Transformer model. The latent representation is then fed into the diffusion model to generate new policy parameters. We implement a task-conditioned latent diffusion model to realize effective parameter generation from simple prompts. We evaluate DPPG on a variety of reinforcement learning tasks including Classic Control suites, Atari games, and MuJoCo locomotion tasks. Our experiments show that DPPG can generate high-quality policy parameters for different reinforcement learning tasks within one pre-trained model. Notably, in up to 20 Atari, 5 Classic Control, and 5 MuJoCo tasks, the policy generated by a unified DPPG model can achieve performance comparable to the original policy trained from scratch in each task. Moreover, we find that DPPG exhibits a certain degree of generalization: it can generate parameters in unseen environments that converge faster and achieve higher performance.

To summarize, our contributions are as follows:
\begin{itemize}
    \item We are the first to propose generating policy parameters for decision-making tasks directly via diffusion models.
    \item We propose DPPG, which innovatively integrates an autoencoder and a diffusion model with fully condition-based Transformer-based modules. This architecture enables the generation of large-scale parameter policy networks and allows training a unified model across multiple tasks.
    \item We show that one pre-trained DPPG model can generate policy parameters for various tasks with performance equal to or even exceeding that of policies trained with RL from scratch.
    \item We observe that DPPG shows decent generalization capabilities: it can generate parameters in previously unseen environments that converge faster and achieve higher performance.
\end{itemize}

\section{Related Work}

\paragraph{Multi-task RL and Meta-RL.} Although reinforcement learning algorithms are commonly developed to train policies in a single environment, recent research has shown that multi-task reinforcement learning can significantly improve learning efficiency by simultaneously training across various tasks \citep{dqn, mtrl, huang2022curriculum, zhang2022dynamics}. Typically, multi-task reinforcement learning is achieved by sharing the policy network across different tasks; however, the network structure cannot be shared when facing different observation spaces and action spaces. To handle this, recent works usually add different output heads after a pre-trained representation extractor, or factorize the policy into conditional components so that a single backbone can serve many agents \citep{yu2022centralized}. Meta-Reinforcement Learning (Meta-RL) represents an evolution within the field, focusing on algorithms that can efficiently adapt to new tasks with minimal data \citep{maml, hypernetworks, meta-rl}, thereby addressing the limitations of traditional RL in terms of data efficiency and policy generality. Transformer-based policy backbones have further unified model design across tasks \citep{li2023transformerrl}, while self-play and population-based training remain key drivers of robust multi-agent behavior \citep{wu2021coordinated, zhang2024selfplay}. Unlike these approaches, which still rely on RL training methods, we attempt to use a diffusion model to directly generate the parameters of the policy network. These generated parameters exhibit performance in different environments comparable to the original policy.

Complex game-control systems complement this line of work: Ye et al. study large-scale reinforcement learning for MOBA action control \citep{ye2020complexcontrol} and combine curriculum self-play with policy distillation to train across a large pool of heroes \citep{ye2020fullmoba}; supervised distillation alone can also reach human-level MOBA play \citep{ye2022juewusl}. Beyond MOBAs, hierarchical RL has been applied to open-world games such as Minecraft \citep{lin2021juewumc}, and real-game datasets now support offline RL benchmarking on commercial titles \citep{wei2022hokarena, qu2024hokoff}. A parallel line of work treats playable games themselves as a generative target \citep{yang2024playable}. More recently, LLM-based agents have shifted the focus toward multi-agent orchestration \citep{ye2024moreagents, yu2024affordable} and world-model alignment \citep{zhou2024walle}. These works motivate scalable and transferable policy learning, whereas our focus is on generating the policy parameters themselves.

\paragraph{Diffusion models.} The research of diffusion models prevails with the introduction of denoising diffusion probabilistic models (DDPM) \citep{ddpm}. Diffusion models are trained with a forward process and a reverse process: the forward process adds noise to the original sample, and the reverse process removes the noise to recover the original sample. Diffusion models have shown remarkable progress in generating high-quality images and videos \citep{stable-diffusion, dalle2, imagen-video}. Since directly performing diffusion processes in the original input space is typically time-inefficient, recent works propose diffusing the latent representation of the input data, which can be more efficient and effective in practice \citep{dit, ddim, analytic-dpm}. In this paper, we adopt the latent diffusion model to generate structural policy parameters for reinforcement learning tasks, which can significantly reduce the compute and memory required compared with directly diffusing original parameters.

\paragraph{Parameter generation.} Parameter generation is a step forward from the field of automated machine learning (AutoML) \citep{automl-survey}, which considers automating the end-to-end process of training machine learning models. Prior works deploy hypernetworks \citep{hypernet, smash} or variational autoencoders \citep{G.pt} to generate network parameters. With the introduction of diffusion models, recent approaches significantly improve the quality of generated parameters with prompt-based and diverse synthesis results \citep{pdiff, d2nwg, gpd, hyperdiffusion}. However, these works on parameter generation mainly focus on generating parameters for supervised learning tasks such as image classification and tabular data prediction. In parallel, the RL community has developed complementary tools for obtaining good policy parameters without training a separate network from scratch: policy adaptation methods reuse a trained backbone and adjust it to new tasks, while robust offline-to-online techniques \citep{gao2022robustoffline, he2025robustpolicy}, discrete actor-critic improvements \citep{zhou2022sac}, variance-reduction tricks in policy optimization \citep{lin2023sampledropout, wu2021coordinated}, and feedback-driven RL \citep{liu2023rltf} all aim at the same goal of producing effective policy parameters more cheaply. In this paper, we extend the parameter generation paradigm to reinforcement learning tasks, where the generated parameters are used to initialize the policy network for different tasks. We show that DPPG can generate high-quality policy parameters for multiple reinforcement learning tasks with only one pre-trained model.

\section{Approach}
\label{Approach}

\subsection{Preliminaries}
Diffusion models, drawing inspiration from the principles of thermodynamic diffusion, operate as sequential models with hidden variables. The training methodology for these models is bifurcated into two distinct phases: the introduction of noise in a forward manner and the subsequent reverse noise reduction. In the initial phase, Gaussian noise is incrementally introduced to the original sample, whereas the latter phase embarks from a state of random noise, employing a Markov chain to facilitate the training of a Gaussian transformation.

\paragraph{Forward Process.}
Given an initial sample $x_0 \sim q(x)$, the forward process incrementally infuses Gaussian noise through $T$ iterative steps, culminating in a series of intermediate representations ($x_1$, $x_2$, \ldots, $x_T$). The formal description of this progression is articulated as follows:
\begin{equation}
    q\left(x_{t} \mid x_{t-1}\right)=\mathcal{N}\left(x_{t} ; \sqrt{1-\beta_{t}} x_{t-1}, \beta_{t} \mathbf{I}\right)
\end{equation}
\begin{small}
\begin{equation}
\begin{split}
    q\left(x_{t} \mid x_{t-1}\right) &= \mathcal{N}\left(x_{t} ; \sqrt{1-\beta_{t}} x_{t-1}, \beta_{t} \mathbf{I}\right) \\
    q\left(x_{1: T} \mid x_{0}\right) &= \prod_{t=1}^{T} q\left(x_{t} \mid x_{t-1}\right)
\end{split}
\end{equation}
\end{small}

Here, $q$ signifies the process of adding noise in the forward direction, while $\mathcal{N}$ stands for the Gaussian noise parameterized by $\beta_t$. The symbol $\mathbf{I}$ denotes the identity matrix.

\paragraph{Reverse Process.}
The reverse denoising process acts as the counterpart to the forward process, systematically eliminating noise from $x_t$ by incrementally training a denoising neural network over $T$ steps to yield the original sample. The iterative equation governing the reverse denoising process is presented as follows:
\begin{equation}
p_{\theta}\left(x_{t-1} \mid x_{t}\right)=\mathcal{N}\left(x_{t-1} ; \mu_{\theta}\left(x_{t}, t\right), \Sigma_{\theta}\left(x_{t}, t\right)\right)
\end{equation}
\begin{equation}
p_{\theta}\left(x_{0: T}\right)=p\left(x_{T}\right) \prod_{t=1}^{T} p_{\theta}\left(x_{t-1} \mid x_{t}\right)
\end{equation}
where $p$ denotes the reverse process, while $\mu_{\theta}(x_t, t)$ and $\Sigma_{\theta}(x_t, t)$ represent the Gaussian mean and variance, respectively, as estimated by the denoising network with parameters $\theta$. The optimization of the denoising network within the reverse process is achieved through minimizing the standard negative log-likelihood:
\begin{equation}
L_{\mathrm{dm}}=\mathcal{D}_{K L}\left(q\left(x_{t-1} \mid x_{t}, x_{0}\right) \| p_{\theta}\left(x_{t-1} \mid x_{t}\right)\right)
\end{equation}
where $D_{KL}(\cdot \| \cdot)$ signifies the Kullback-Leibler (KL) divergence, a measure commonly employed to quantify the discrepancy between two probability distributions.

\subsection{Overall Framework}
\begin{figure}[!htbp]
  \centering
  \includegraphics[width=0.9\textwidth]{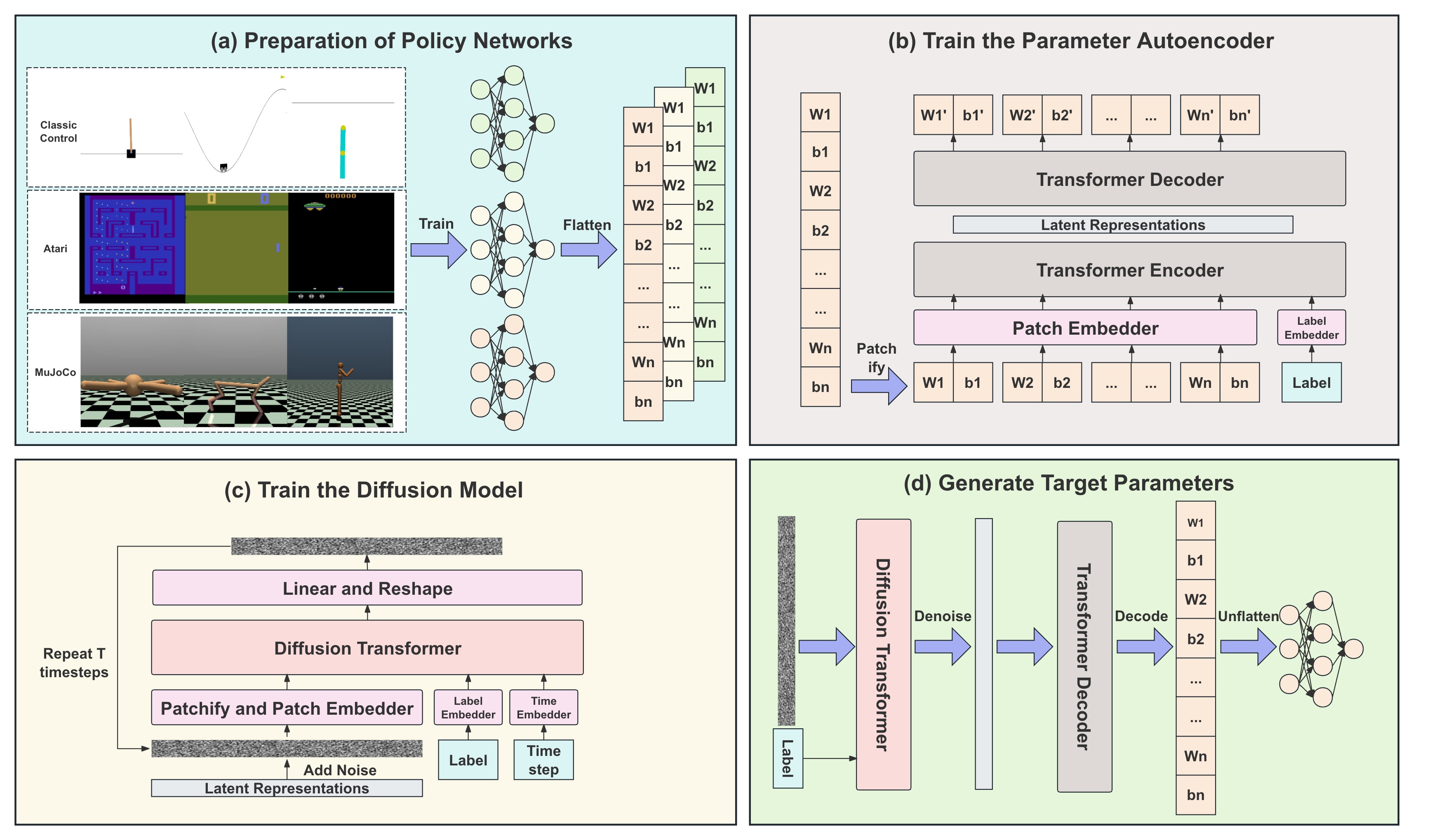}
  \caption{An overview of the proposed framework. (a) Prepare the parameters of policy networks trained in various reinforcement learning environments.
(b) Train a Transformer-based autoencoder to compress the parameters.
(c) Train a diffusion model to generate the compressed latent representations.
(d) Reconstruct the final policy network parameters from random noise during inference.}
  \label{fig_framework}
\end{figure}

Figure~\ref{fig_framework} presents the overall structure of our proposed framework, which is divided into four key components. First, we employ reinforcement learning to train policy models for various environments, resulting in a collection of optimized policy network parameters. Second, since directly generating the entire network's parameters is challenging, we train an autoencoder to compress the parameters. Third, we utilize the latent representation of these compressed parameters to train a diffusion model. Fourth, during inference, the final policy network's parameters are reconstructed from random noise. Detailed explanations of these phases are provided in the following subsections.

\subsection{Preparation of Policy Networks}
To begin, we trained a corresponding policy network for each common reinforcement learning environment, including but not limited to Classic Control, Atari, and MuJoCo. Due to the differences in observation space, action space, and the complexity of the environments, the structure and size of these policy networks vary significantly. For instance, networks for the Classic Control environment primarily consist of MLP layers with thousands of parameters. In contrast, networks designed for the MuJoCo environment are composed of MLP layers with tens of thousands of parameters. The Atari environment, being a video game setting, is considerably more complex, featuring networks with millions of parameters that include both CNN and MLP layers. Notably, our method is insensitive to the model's structure and size; it is capable of generating network parameters for any architecture.

Upon completing the training of policy networks for each reinforcement learning environment, we proceed to fine-tune each environment's policy for $n \times K$ steps. During this fine-tuning process, a checkpoint is saved every $n$ steps, resulting in a total of $K$ checkpoints. These checkpoints' parameters are then utilized as our training data. In our experiments, we typically set $n \times K$ equal to the number of steps taken to initially converge the policy network, ensuring the diversity of the input data. For each environment, we finally obtain a parameter dataset $\Theta=\{\theta_1,\theta_2,...,\theta_K\}$ to train the autoencoder.

\subsection{Train the Parameter Autoencoder}
\paragraph{Patchify.}
Following this, we flatten all parameters into 1-dimensional vectors $V=\{v_1,v_2,...,v_K\}$ and segment these vectors into patches, each containing $p$ parameters, arranged in an order from shallow to deep layers, as illustrated in Figure~\ref{fig_framework}(b). Here, $p$ represents the patch size.
Subsequently, a patch embedding layer is utilized to transform the patchified parameters into tokens that can be fed into a Transformer model. This patchification approach ensures our method's capability to handle input parameters of any size and structure.

\paragraph{Transformer-based Autoencoder.}
Then, we feed the transformed patch tokens into a Transformer-based autoencoder, which maintains encoder and decoder structures identical to those of a standard Transformer. To further enhance the reconstruction capability of the autoencoder, we encode the environment index as a label and treat it as an additional token input to the Transformer encoder. Moreover, to bolster the robustness and generalization capacity of the autoencoder, we follow the work of \cite{pdiff} and implement random noise augmentation on both the input parameters and the latent representations. In alignment with conventional autoencoder training practices, we aim to minimize the mean squared error (MSE) loss between $V$ and $V'$ as follows:

\begin{equation}
L_{{\mathrm{MSE}}}=\frac1K\sum_1^K\lVert v_k-v_k^{\prime}\rVert^2
\end{equation}

where $v_k^{\prime}$ is the $k$-th reconstructed parameter.

Ultimately, we obtain the latent representation $z$ of the vector $v$, where the size of $z$ is $[\text{patch\_num}, d\_model]$. Here, $\text{patch\_num}$ refers to the number of patches after patchifying $v$, and $d\_model$ represents the dimension of each patch after compression.

\subsection{Train the Diffusion Model}
Due to the challenges associated with directly generating the full set of parameters, we follow the approach outlined in \cite{stable-diffusion} and utilize the latent representation $z$, obtained from the autoencoder, to train a diffusion model. This model is conditioned on the environment index $i$ and the current timestep $t$, as illustrated in Figure~\ref{fig_framework}(c). In our experiments, we flatten $z$ and then re-patchify it, typically selecting a divisor of $d\_model$ as the patch size to avoid disrupting the structure of $z$. Ultimately, we optimize the following latent diffusion model objective:
\begin{equation}\mathcal{L}_{LDM}=\mathbb{E}_{z,\varepsilon\sim\mathcal{N}(0,1),i,t}\left[||\varepsilon-\varepsilon_\psi(z_t,i,t)||_2^2\right]\end{equation}
where $\varepsilon$ represents the added Gaussian noise, and $\varepsilon_\psi(z_t, i, t)$ denotes the denoising network parameterized by $\psi$. Here, we opt for the Diffusion Transformer \citep{dit} as the denoising network, which we adapt to suit the diffusion of one-dimensional data, while striving to maintain as much of its original setting as possible.

\subsection{Generate Target Parameters}
As illustrated in Figure~\ref{fig_framework}(d), new parameters are generated conditionally through the reverse diffusion process as follows:
\begin{equation}
    z_t=\frac1{\sqrt{a_t}}(z_t-\frac{\beta_t}{\sqrt{1-\tilde{a}_t}}\varepsilon_\psi(z_t,i,t))+\sigma\xi
\end{equation}
where $\xi\sim\mathcal{N}(0,I)$. After generating a latent representation $\bar{z}$ for environment $E_i$, we utilize the AE decoder to reconstruct the vector $\bar{v}$, i.e., $\bar{v}=\text{Decoder}(\bar{z})$.
Finally, we unflatten $\bar{v}$ to obtain the final target parameters $\bar{\theta}$.

\section{Experiments}
\label{Experiments}

\paragraph{Testing Environments.}
We selected Atari games \citep{atari}, Classic Control \citep{gym}, and MuJoCo \citep{mujoco} as our test environments due to their widespread acceptance in the reinforcement learning community as benchmarks for evaluating our method.

\paragraph{Setting.}
For each sub-environment, we initially train an agent using Proximal Policy Optimization (PPO) \citep{ppo} until convergence. Subsequently, we fine-tune this model to obtain parameters from 200 checkpoints, which serve as our training dataset. The architecture of our autoencoder consists of a three-layer Transformer encoder and a three-layer Transformer decoder.

\paragraph{RL Zoo.}
We utilize the "RL Baselines3 Zoo" project as the source for our experiment's initial hyperparameters. This project is designed to facilitate the training, evaluation, and fine-tuning of reinforcement learning agents across a wide array of environments. It builds upon the Stable Baselines3 library, which is a set of reliable implementations of state-of-the-art RL algorithms in PyTorch. The RL Baselines3 Zoo extends this by providing a curated collection of pre-trained models and hyperparameters that have been rigorously optimized for performance across various tasks.

\paragraph{Baselines.}
We selected the returns of the original Proximal Policy Optimization (PPO) \citep{ppo} algorithm and pdiff \citep{pdiff} in the respective environments as our baseline.

\subsection{Performance in Single Environments}
Table~\ref{table_single} shows the performance in single environments. This means that we trained the DPPG network within individual environments and conducted tests within those same environments. For Classic Control and MuJoCo tasks, we directly generate the parameters for the entire network. For Atari tasks, given the larger size of the network, we generate parameters for the action layer. The situation of generating parameters for large networks will be discussed in Section~\ref{sec:large_parameters}. The results indicate that DPPG can generate superior parameters across 20 Atari tasks, 5 Classic Control tasks, and 5 MuJoCo tasks. In the vast majority of these environments, the parameters generated by our method outperform those trained by the original PPO algorithms. This suggests that our approach is capable of learning the parameter distribution of the initial policy network and generating improved parameters from random noise.
Additionally, the performance generated by our method is generally better than that of pdiff, and the variance in returns is smaller, which demonstrates the stability of our approach. Notably, pdiff encountered an Out of Memory (OOM) issue when generating parameters for a network in the Humanoid environment, which consists of 166,673 parameters, while our method did not. This indicates that our method can generate larger parameters using less GPU memory. Detailed discussions on this will also be presented in Section~\ref{sec:large_parameters}.

\begin{table}[!htbp]
\centering
\small{
\caption{\textbf{Performance in Single Environments.} We train DPPG within a single environment and test within the same setting. "Parameter Number" refers to the size of the network parameters. "Original" denotes the returns of the baseline RL model. For each environment, we generate 200 sets of network parameters and report the mean and variance of their returns.}
\begin{tabular}{@{}llclll@{}}
\toprule
 & Environment & \makecell{Parameter\\number} & Original & pdiff & DPPG (Ours) \\ \midrule
\multirow{20}{*}{Atari}
& Alien & 9234 & $674.78_{\pm 48.17}$ & \bm{$856.55_{\pm 238.60}$} & $735.0_{\pm 50.69}$ \\
 & Amidar & 5130 & $761.22_{\pm 12.34}$ & $583.51_{\pm 220.35}$ & \bm{$772.68_{\pm 14.94}$} \\
 & Assault & 3591 & $5562.01_{\pm 739.71}$ & $153.93_{\pm 401.77}$ & \bm{$6500.79_{\pm 638.09}$} \\
 & Asteroids & 7182 & $1160.70_{\pm 99.18}$ & $865.62_{\pm 307.34}$ & \bm{$1185.55_{\pm 97.08}$} \\
 & BattleZone & 9234 & $23556.0_{\pm 1743.23}$ & $23635.0_{\pm 4832.83}$ & \bm{$28113.0_{\pm 2029.09}$} \\
 & BeamRider & 4617 & $4183.13_{\pm 455.50}$ & $4177.47_{\pm 828.9}$ & \bm{$4364.60_{\pm 517.49}$} \\
 & Boxing & 9234 & $97.41_{\pm 1.41}$ & $78.49_{\pm 38.00}$ & \bm{$97.60_{\pm 1.61}$} \\
 & Breakout & 2052 & \bm{$478.99_{\pm 53.59}$} & $322.05_{\pm 77.95}$ & $394.61_{\pm 7.33}$ \\
 & CrazyClimber & 4617 & \bm{$136518.3_{\pm 5195.24}$} & $92272.5_{\pm 25370.07}$ & $129752.55_{\pm 6395.88}$ \\
 & Enduro & 4617 & \bm{$911.96_{\pm 53.20}$} & $765.33_{\pm 201.31}$ & $904.85_{\pm 62.86}$ \\
 & Freeway & 1539 & $32.88_{\pm 0.13}$ & $32.88_{\pm 0.14}$ & \bm{$32.90_{\pm 0.13}$} \\
 & Jamesbond & 9234 & $1282.35_{\pm 463.58}$ & $1142.1_{\pm 668.88}$ & \bm{$1311.95_{\pm 489.22}$} \\
 & Kangaroo & 9234 & $1800_{\pm 0}$ & $1630.5_{\pm 390.11}$ & \bm{$1800_{\pm 0}$} \\
 & MsPacman & 4617 & $2018.1_{\pm 271.45}$ & $2175.6_{\pm 338.37}$ & \bm{$2309.83_{\pm 157.64}$} \\
 & Pong & 3078 & $20.41_{\pm 0.23}$ & $18.19_{\pm 6.99}$ & \bm{$20.45_{\pm 0.26}$} \\
 & Qbert & 3078 & $14819.37_{\pm 838.42}$ & $14742.5_{\pm 2514.34}$ & \bm{$15092.87_{\pm 819.51}$} \\
 & RoadRunner & 9234 & $27458.85_{\pm 2137.92}$ & $25228.5_{\pm 4499.96}$ & \bm{$28372.0_{\pm 2670.89}$} \\
 & Robotank & 9234 & $25.43_{\pm 2.47}$ & $18.86_{\pm 2.74}$ & \bm{$26.05_{\pm 2.26}$} \\
 & Seaquest & 9234 & $1796.13_{\pm 9.83}$ & $1704.9_{\pm 281.24}$ & \bm{$1802.99_{\pm 11.15}$} \\
 & SpaceInvaders & 3078 & $884.80_{\pm 133.86}$ & $849.05_{\pm 237.46}$ & \bm{$976.64_{\pm 174.12}$} \\
 \midrule
\multirow{5}{*}{\makecell{Classic-\\Control}} & CartPole & 4610 & $500_{\pm 0}$ & $500_{\pm 0}$ & \bm{$500_{\pm 0}$} \\
 & Acrobot & 4803 & $-67.55_{\pm 4.42}$ & $-100.30_{\pm 93.48}$ & \bm{$-67.39_{\pm 3.64}$} \\
 & MountainCar & 4547 & $-116.56_{\pm 7.65}$ & $-121.97_{\pm 19.46}$ & \bm{$-114.53_{\pm 6.57}$} \\
 & \makecell{MountainCarContinuous} & 4417 & $46.32_{\pm 71.37}$ & $60.34_{\pm 61.82}$ & \bm{$93.02_{\pm 1.80}$} \\
 & Pendulum & 4481 & $-171.18_{\pm 37.15}$ & $-380.98_{\pm 475.25}$ & \bm{$-152.13_{\pm 35.51}$} \\
 \midrule
\multirow{5}{*}{MuJoCo} & Ant & 6472 & $996.03_{\pm 715.68}$ & $660.47_{\pm 902.87}$ & \bm{$1067.14_{\pm 413.51}$} \\
 & HalfCheetah & 71942 & $5079.27_{\pm 1314.46}$ & $2576.34_{\pm 1129.98}$ & \bm{$5920.20_{\pm 65.78}$} \\
 & Hopper & 69635 & $961.03_{\pm 117.21}$ & $840.5_{\pm 101.48}$ & \bm{$1005.59_{\pm 0.62}$} \\
 & Humanoid & 166673 & $705.68_{\pm 157.74}$ & OOM & \bm{$874.59_{\pm 73.24}$} \\
 & Walker2d & 5702 & $2703.18_{\pm 706.58}$ & $1711.36_{\pm 1246.77}$ & \bm{$3124.33_{\pm 295.78}$} \\
\bottomrule
\end{tabular}
\label{table_single}
}
\end{table}

\subsection{Performance Across the Entire Environments}
Table~\ref{table_entire} shows the performance across the entire environments. In this setup, we trained a unified DPPG across all environments and generate parameters for the corresponding environment from random noise and the environment's label. Since pdiff can only be trained in a single environment and generate parameters for that specific environment, we only compare its performance with the original RL model. It can be observed that, although the multi-task learning aspect may result in some tasks not performing as well as in the single environment setting, in most tasks, DPPG is still capable of generating parameters that yield higher returns than those of the original RL model.

\begin{table}[!htbp]
\centering
\small{
\caption{\textbf{Performance Across the Entire Environments.} We trained a unified DPPG model across all environments and conducted tests within their respective settings. "Parameter Number" refers to the size of the network parameters. "Original" denotes the returns of the baseline RL model. For each environment, we generate 200 sets of network parameters and report the mean and variance of their returns.}
\begin{tabular}{@{}llcll@{}}
\toprule
 & Environment & Parameter number & Original & DPPG (Ours) \\
\midrule
\multirow{20}{*}{Atari} & Alien & 9234 & $674.78_{\pm 48.17}$ & \bm{$727.14_{\pm 41.18}$} \\
 & Amidar & 5130 & $761.22_{\pm 12.34}$ & \bm{$761.79_{\pm 13.27}$} \\
 & Assault & 3591 & \bm{$5562.01_{\pm 739.71}$} & $5140.74_{\pm 654.34}$ \\
 & Asteroids & 7182 & $1160.70_{\pm 99.18}$ & \bm{$1208.31_{\pm 105.47}$} \\
 & BattleZone & 9234 & \bm{$23556.0_{\pm 1743.23}$} & $21253.5_{\pm 2143.00}$ \\
 & BeamRider & 4617 & $4183.13_{\pm 455.50}$ & \bm{$4511.57_{\pm 414.10}$} \\
 & Boxing & 9234 & \bm{$97.41_{\pm 1.41}$} & $96.58_{\pm 1.50}$ \\
 & Breakout & 2052 & \bm{$478.99_{\pm 53.59}$} & $376.38_{\pm 53.59}$ \\
 & CrazyClimber & 4617 & \bm{$136518.3_{\pm 5195.24}$} & $75424.55_{\pm 679.24}$ \\
 & Enduro & 4617 & \bm{$911.96_{\pm 53.20}$} & $894.49_{\pm 47.57}$ \\
 & Freeway & 1539 & \bm{$32.88_{\pm 0.13}$} & $32.67_{\pm 0.19}$ \\
 & Jamesbond & 9234 & \bm{$1282.35_{\pm 463.58}$} & $1275.25_{\pm 403.93}$ \\
 & Kangaroo & 9234 & $1800_{\pm 0}$ & \bm{$1800_{\pm 0}$} \\
 & MsPacman & 4617 & \bm{$2018.1_{\pm 271.45}$} & $1483.27_{\pm 65.14}$ \\
 & Pong & 3078 & $20.41_{\pm 0.23}$ & \bm{$20.42_{\pm 0.19}$} \\
 & Qbert & 3078 & \bm{$14819.37_{\pm 838.42}$} & $14683.02_{\pm 838.77}$ \\
 & RoadRunner & 9234 & $27458.85_{\pm 2137.92}$ & \bm{$28018.0_{\pm 1778.81}$} \\
 & Robotank & 9234 & \bm{$25.43_{\pm 2.47}$} & $24.77_{\pm 2.18}$ \\
 & Seaquest & 9234 & $1796.13_{\pm 9.83}$ & \bm{$1800.43_{\pm 7.24}$} \\
 & SpaceInvaders & 3078 & $884.80_{\pm 133.86}$ & \bm{$941.86_{\pm 116.01}$} \\
\midrule
\multirow{5}{*}{\makecell{Classic-\\Control}} & CartPole & 4610 & $500_{\pm 0}$ & \bm{$500_{\pm 0}$} \\
 & Acrobot & 4803 & $-67.55_{\pm 4.42}$ & \bm{$-66.9_{\pm 3.12}$} \\
 & MountainCar & 4547 & \bm{$-116.56_{\pm 7.65}$} & $-122.46_{\pm 6.96}$ \\
 & MountainCarContinuous & 4417 & $46.32_{\pm 71.37}$ & \bm{$91.69_{\pm 0.67}$} \\
 & Pendulum & 4481 & $-171.18_{\pm 37.15}$ & \bm{$-170.95_{\pm 29.39}$} \\
\midrule
\multirow{5}{*}{MuJoCo} & Ant & 6472 & \bm{$996.03_{\pm 715.68}$} & $504.54_{\pm 140.64}$ \\
 & HalfCheetah & 71942 & \bm{$5079.27_{\pm 1314.46}$} & $1949.26_{\pm 69.53}$ \\
 & Hopper & 69635 & $961.03_{\pm 117.21}$ & \bm{$1009.37_{\pm 0.29}$} \\
 & Humanoid & 166673 & \bm{$705.68_{\pm 157.74}$} & $436.35_{\pm 37.54}$ \\
 & Walker2d & 5702 & $2703.18_{\pm 706.58}$ & \bm{$3030.85_{\pm 265.42}$} \\
\bottomrule
\end{tabular}
\label{table_entire}
}
\end{table}

\subsection{Performance on Large Parameters}
\label{sec:large_parameters}
The capacity to generate a large number of parameters is also a crucial metric. Previous work \citep{pdiff, d2nwg, gpd, hyperdiffusion} has predominantly tested on networks with relatively small parameter sizes.
Table~\ref{table_large} presents the performance results on large parameter networks using Atari's 20 games as the testing environment. These environments have network parameter sizes of approximately 1.69 million. Due to the Out-Of-Memory (OOM) issue with pdiff at this parameter scale, we only compare the performance with the original RL model. Thanks to our Transformer-based autoencoder's ability to achieve extremely high compression rates, DPPG is capable of generating competitive parameters even in high-parameter environments.

\begin{figure}[!htbp]
\centering
\includegraphics[width=0.46\linewidth]{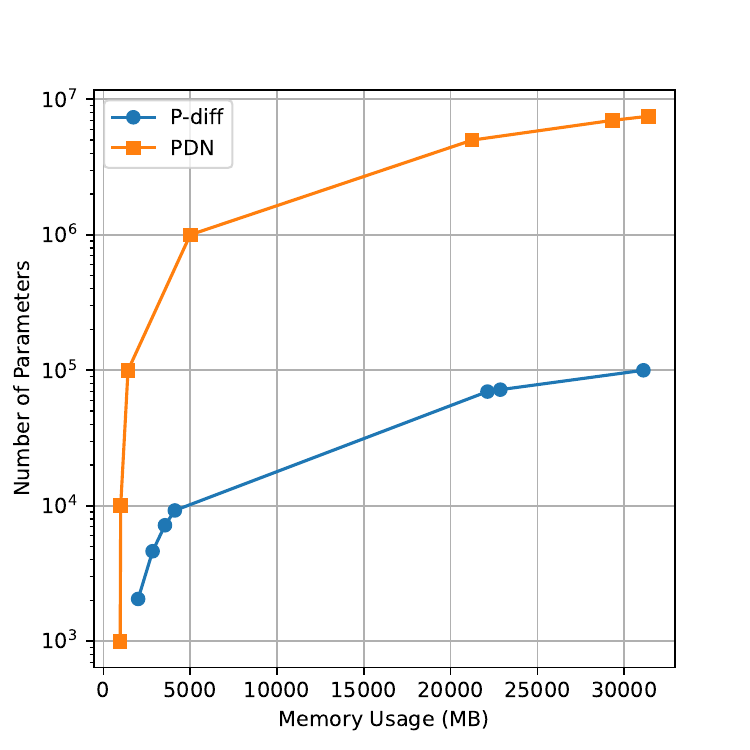}
\caption{Under the same GPU memory, DPPG can generate a network with approximately 100 times the number of parameters as pdiff.}
\label{fig_memory}
\end{figure}

Figure~\ref{fig_memory} illustrates the relationship between the required GPU memory and the number of parameters generated for DPPG and pdiff, with a batch size of 200. It is clear that under the same GPU memory constraints, DPPG can generate a network with approximately 100 times the number of parameters compared to pdiff.

\begin{table}[!htbp]
\centering
\small{
\caption{\textbf{Performance on Large Parameters.} "Parameter Number" refers to the size of the network parameters. "Original" denotes the returns of the baseline RL model. For each environment, we generate 200 sets of network parameters and report the mean and variance of their returns.}
\begin{tabular}{@{}llcll@{}}
\toprule
 & Environment & Parameter number & Original & DPPG (Ours) \\
\midrule
\multirow{20}{*}{Atari}
    & Alien & 1693362 & $763.28_{\pm 82.81}$ & \bm{$799.75_{\pm 98.65}$} \\
    & Amidar & 1684641 & $724.78_{\pm 19.60}$ & \bm{$763.36_{\pm 20.66}$} \\
    & Assault & 1683102 & $5270.33_{\pm 696.26}$ & \bm{$5688.04_{\pm 514.72}$} \\
    & Asteroids & 1686693 & \bm{$1198.2_{\pm 112.04}$} & $1189.4_{\pm 83.49}$ \\
    & BattleZone & 1693362 & $19131.5_{\pm 1968.21}$ & \bm{$20725.0_{\pm 2310.16}$} \\
    & BeamRider & 1688745 & $4437.67_{\pm 606.95}$ & \bm{$4548.55_{\pm 477.68}$} \\
    & Boxing & 1693362 & \bm{$97.58_{\pm 1.15}$} & $97.47_{\pm 0.99}$ \\
     & Breakout & 1681563 & \bm{$406.22_{\pm 42.99}$} & $381.51_{\pm 19.73}$ \\
     & CrazyClimber & 1688745 & $119060.5_{\pm 8300.94}$ & \bm{$120772.5_{\pm 5890.09}$} \\
     & Enduro & 1688745 & \bm{$915.46_{\pm 58.55}$} & $849.10_{\pm 43.76}$ \\
     & Freeway & 1681050 & \bm{$32.74_{\pm 0.12}$} & $32.36_{\pm 0.27}$ \\
     & Jamesbond & 1693362 & $1247.0_{\pm 403.93}$ & \bm{$1356.0_{\pm 499.48}$} \\
     & Kangaroo & 1693362 & $1800_{\pm 0}$ & \bm{$1800_{\pm 0}$} \\
    & MsPacman & 1688745 & $1936.48_{\pm 241.39}$ & \bm{$2604.95_{\pm 225.12}$} \\
    & Pong & 1687206 & $20.39_{\pm 0.26}$ & \bm{$20.71_{\pm 0.11}$} \\
    & Qbert & 1687206 & \bm{$15513.83_{\pm 1234.25}$} & $14459.37_{\pm 1197.40}$ \\
     & RoadRunner & 1693362 & $27978.7_{\pm 1902.90}$ & \bm{$28561.0_{\pm 2349.40}$} \\
     & Robotank & 1693362 & $24.81_{\pm 2.39}$ & \bm{$25.00_{\pm 1.91}$} \\
    & Seaquest & 1693362 & $1799.9_{\pm 7.73}$ & \bm{$1812.9_{\pm 31.51}$} \\
    & SpaceInvaders & 1687206 & $751.06_{\pm 128.54}$ & \bm{$878.95_{\pm 156.53}$} \\

\bottomrule
\end{tabular}
\label{table_large}
}
\end{table}

\FloatBarrier
\subsection{Generalization Capabilities}
In the previous experiments, we utilized a diffusion model to generate parameters for seen environments. However, generalization is also a crucial capability. In this section, we aim to explore the generalization capabilities of DPPG. To achieve this goal, inspired by the approach in \cite{d2nwg} where images are used as conditions for generating parameters in CV tasks, we use the trajectories of expert gameplay as new conditions.
Specifically, we collected a series of expert trajectories from 20 Atari games using the pre-trained PPO models. These trajectories were then used as data to train a trajectory VAE, which compresses them into a latent space. We replaced the label embedder in Figure~\ref{fig_framework} with this trajectory VAE as the new condition and retrained our DPPG model. Finally, we tested the performance in unseen Atari games.

\begin{figure}[!htbp]
  \centering
  \begin{subfigure}[b]{0.24\textwidth}
    \includegraphics[width=\textwidth]{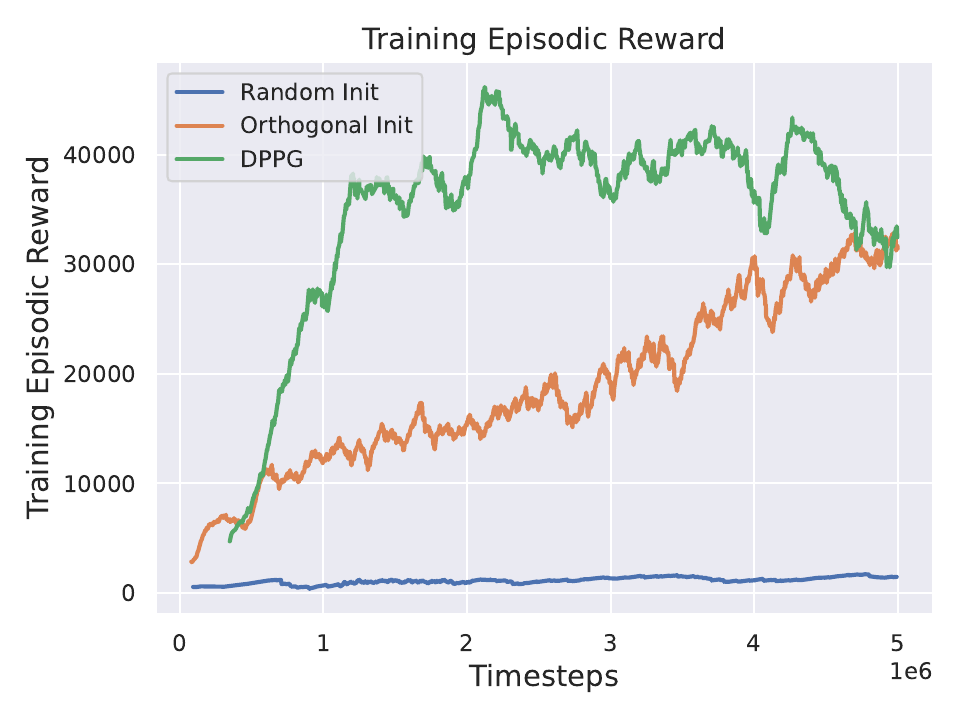}
    \caption{Defender.}
    \label{fig:Defender}
  \end{subfigure}
  \hfill
  \begin{subfigure}[b]{0.24\textwidth}
    \includegraphics[width=\textwidth]{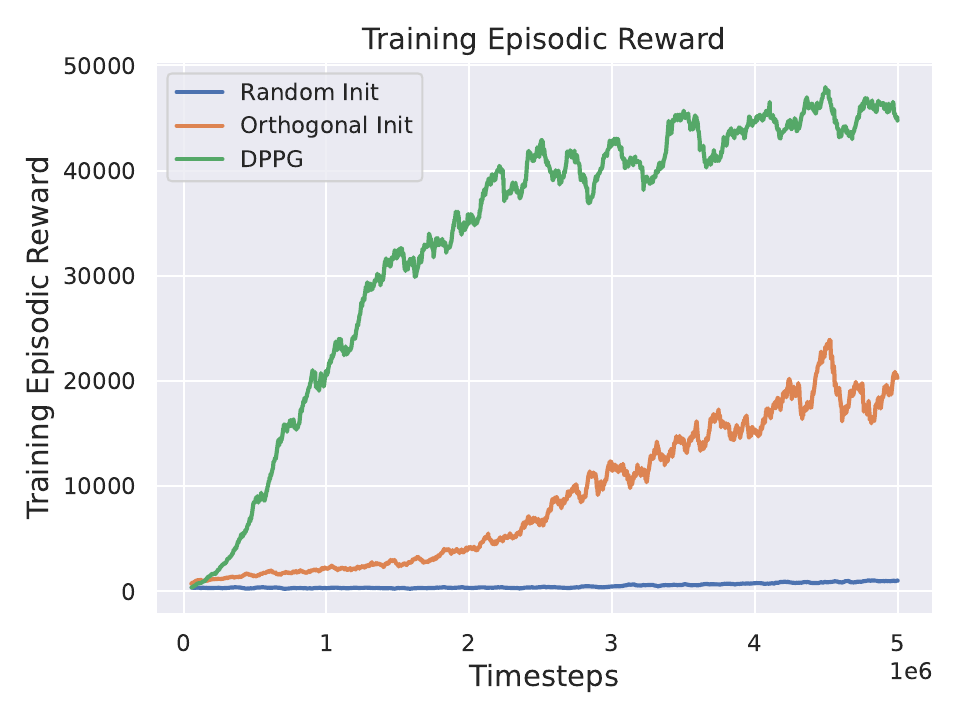}
    \caption{StarGunner.}
    \label{fig:StarGunner}
  \end{subfigure}
    \hfill
  \begin{subfigure}[b]{0.24\textwidth}
    \includegraphics[width=\textwidth]{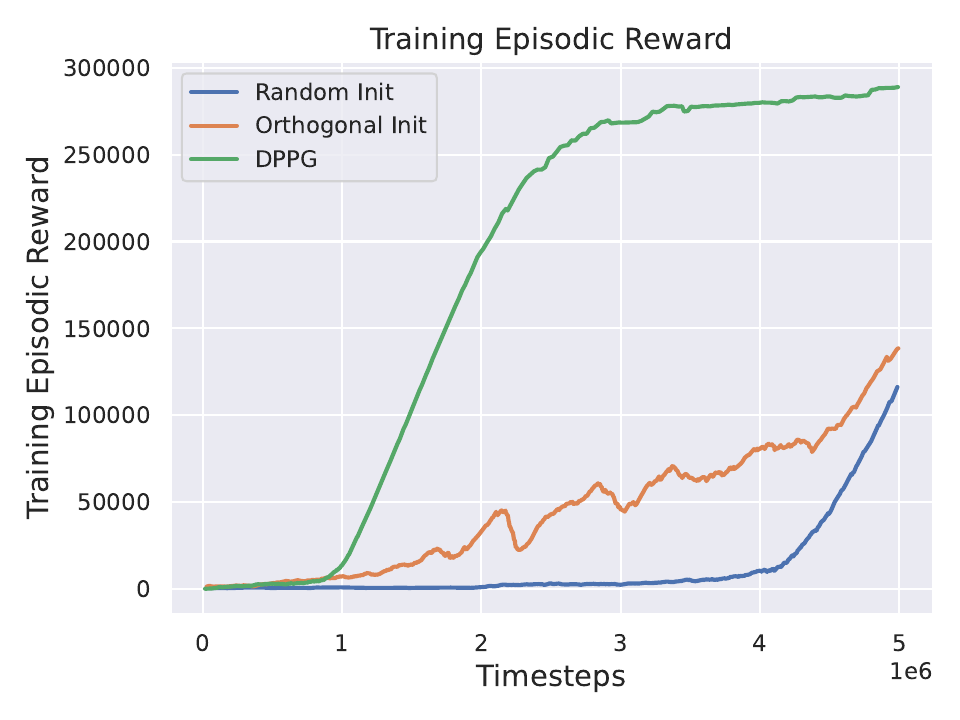}
    \caption{UpNDown.}
    \label{fig:UpNDown}
  \end{subfigure}
    \hfill
  \begin{subfigure}[b]{0.24\textwidth}
    \includegraphics[width=\textwidth]{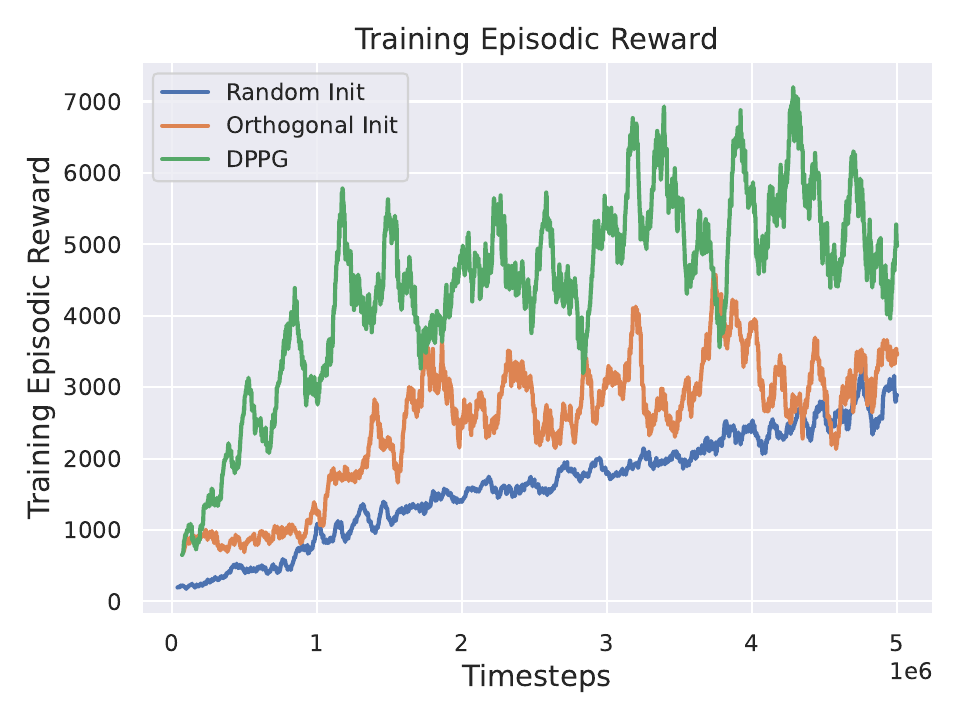}
    \caption{WizardOfWor.}
    \label{fig:WizardOfWor}
  \end{subfigure}

  \caption{Episode reward curve over fine-tuning steps in unseen environments, using generated parameters as initialization.}
  \label{fig:combined_unseen}
\end{figure}

The results are shown in Table~\ref{tab:generation} and Figure~\ref{fig:combined_unseen}. Due to limitations in the quantity and distribution of the data, it is challenging for DPPG to directly generate high-quality policies. Therefore, we evaluated the generalization capabilities by using the parameters generated by DPPG as initialization and, after fine-tuning, comparing these with random initialization and the default orthogonal initialization of PPO from the RL Zoo as baselines.
As observed, although DPPG did not show an advantage when used directly as initial parameters, after fine-tuning, DPPG achieved better performance across all environments and significantly accelerated RL training. This indicates that DPPG can learn the differences in strategies between different environments from the trajectories of similar Atari games and generate reasonable parameters that are closer to the optimal policy.

In the future, we will explore how to directly generate high-quality policy network parameters without the need for any fine-tuning.

\begin{table}[!htbp]
\centering
\caption{\textbf{Performance on Unseen Environments.} We report two sets of results: (a) performance without fine-tuning; (b) performance after fine-tuning.}
\begin{tabular}{lcccccc}
\toprule
Environment & \multicolumn{3}{c}{Initialization} & \multicolumn{3}{c}{Fine-tuning} \\
\cmidrule(r){2-4} \cmidrule(r){5-7}
& Random Init & Orthogonal Init & DPPG & Random Init & Orthogonal Init & DPPG \\
\midrule
AirRaid & $0$ & \bm{$297.5$} & $26.5$ & 0 & 6080 & \bm{$11150$} \\
Defender & $550$ & $1165$ & \bm{$1953$} & 1180 & 30080& \bm{$31190$} \\
DoubleDunk & $-4.0$ & $-4.2$ & \bm{$-3.15$} & -2.8 & -2.4 & \bm{$-0.4$} \\
Gopher & $48$ & $0$ & \bm{$59$} & 1300 & 924 & \bm{$1520$} \\
Phoenix & $58$ & \bm{$612$} & $74.4$ & 3934 & 4888 & \bm{$6524$} \\
StarGunner & $240$ & $150$ & \bm{$290$} & 600 & 21020 & \bm{$49220$} \\
UpNDown & $739$ & \bm{$1520$} & $405$ & 143790 & 203340 & \bm{$289620$} \\
WizardOfWor & $250$ & \bm{$430$} & $117$ & 2420 & 2900 & \bm{$5600$} \\

\bottomrule
\end{tabular}
\label{tab:generation}
\end{table}

\FloatBarrier
\subsection{Ablation Study}
\paragraph{Impact of different autoencoder patch sizes.} The patch size of the autoencoder decides how many patches the original parameters should be divided into and processed to a fixed size of latent representations. In this experiment, we evaluate the policy performance under different autoencoder patch sizes.
Table~\ref{tab_ae_patchsize} shows the effect of different autoencoder patch sizes on generation performance. We use the performance score as our metric, defined as the ratio of the performance of the generated parameters to the performance of the original RL model. It is observed that across different environments, there is a trend of improved generation performance with increasing patch size.
To further investigate this phenomenon, we also experimented with different patch sizes in high-parameter environments.
Surprisingly, we find that our autoencoder can maintain a high compression rate when transforming the input parameter patch sequence into latent representations.
Further analysis of the large-parameter results for Assault and SpaceInvaders is provided in the Appendix (Figure~\ref{fig:combined}).

\begin{table}[!htbp]
\centering
\caption{Average scores with different autoencoder patch sizes}
\begin{tabular}{@{}lcccccc@{}}
\toprule
Environment & 128 & 256 & 512 & 1024 & 2048 \\
\midrule
MountainCar & $0.98$ & $1.03$ & $1.01$ & $1.02$ & $1.02$ \\
Ant & $0.75$ & $0.85$ & $1.10$ & $1.5$ & $1.5$ \\
Assault & $1.09$ & $1.04$ & $1.13$ & $1.15$ & $1.17$ \\
\bottomrule
\end{tabular}
\label{tab_ae_patchsize}
\end{table}

\begin{figure}[!htbp]
\centering
\includegraphics[width=0.37\linewidth]{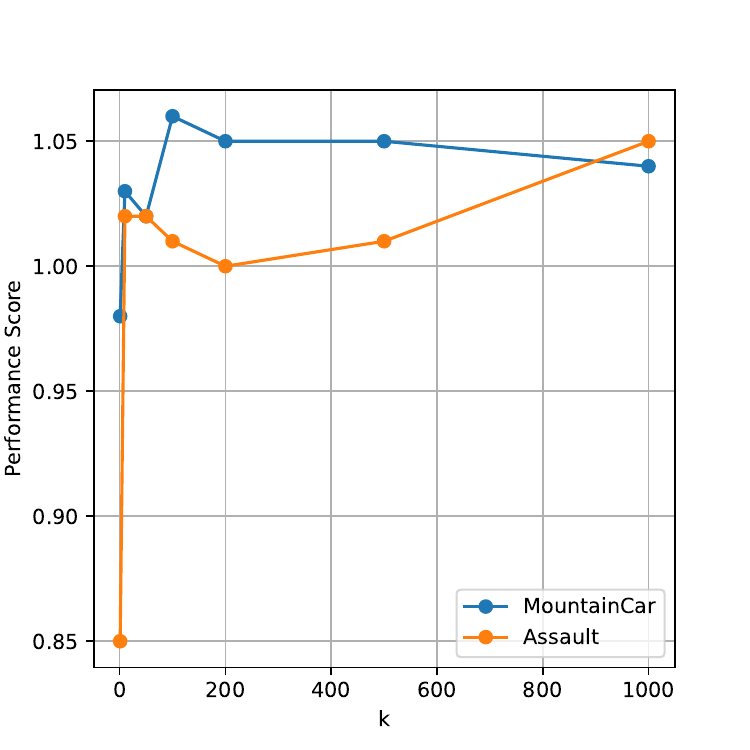}
\caption{Impact of different $K$. $K$ denotes the size of the training data. Increasing $K$ leads to more stable generation outcomes.}
\label{fig_k}
\end{figure}

\paragraph{Impact of different $K$.}
Figure~\ref{fig_k} illustrates the impact of varying $K$ on the results, where $K$ represents the size of the initial training data. The performance score is defined as the ratio of the performance of the generated parameters to the performance of the original RL model. It is evident that as $K$ increases, the generated results become more stable. This aligns with the principle of diffusion models: modeling the initial distribution accurately becomes more challenging with smaller training data.

\FloatBarrier

\section{Conclusion} \label{sec:conclusion}
In this paper, we proposed Diffusion-based Conditional Policy Network Parameter Generation (DPPG), which consists of a Transformer-based autoencoder and a task-conditioned latent diffusion model. Our experiments show that, with one unified DPPG model, we can synthesize policies in a diverse set of tasks from different environment domains including Classic Control, Atari, and MuJoCo.
Additionally, DPPG demonstrates the capability to generate large-parameter policies and exhibits a certain degree of generalization ability.

Although DPPG is capable of generating parameters for unseen environments, the quality of the generated parameters is still constrained by the quantity and distribution of the data. In the future, we will explore methods to directly generate parameters for unseen environments without the need for fine-tuning. Additionally, training a diffusion model capable of generating large-parameter sets remains computationally expensive. We will investigate more efficient designs to address this challenge.

\FloatBarrier
\bibliographystyle{plainnat}
\bibliography{reference}

\clearpage
\appendix
\section*{Appendix}
\phantomsection
\addcontentsline{toc}{section}{Appendix}
\label{sec:appendix}

\subsection*{Impact of different autoencoder patch sizes on large parameters}
\begin{figure}[!htbp]
  \centering
  \begin{subfigure}[b]{0.47\textwidth}
    \includegraphics[width=\textwidth]{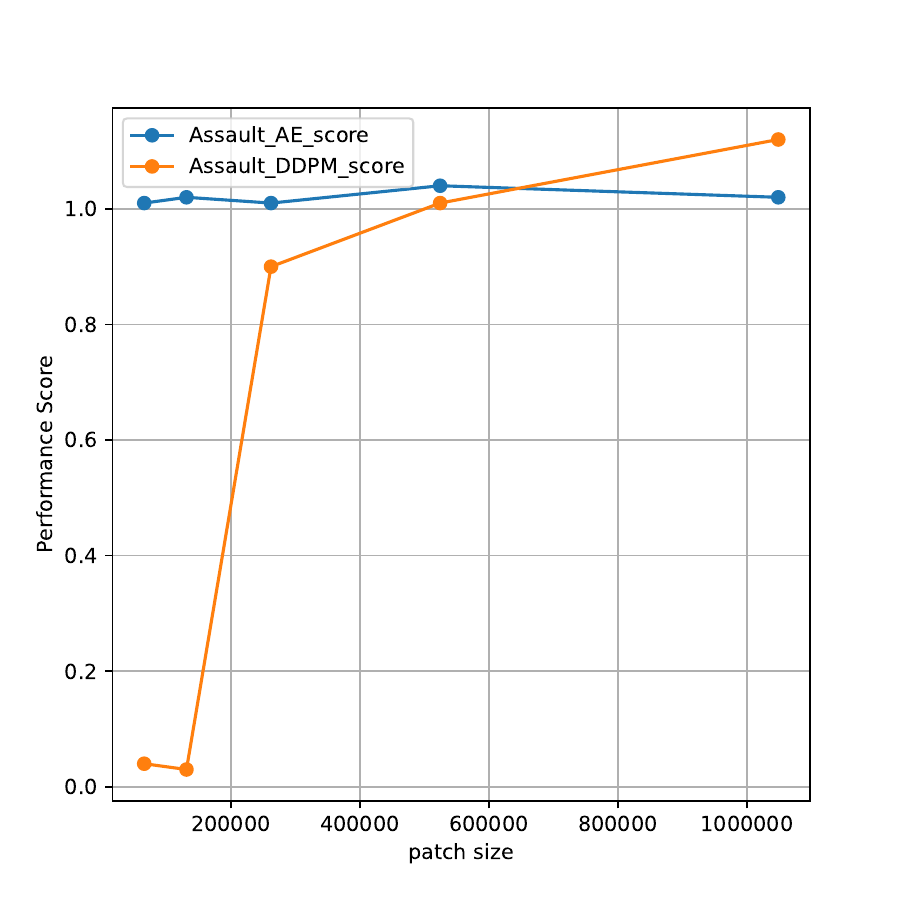}
    \caption{Assault.}
    \label{fig:assault}
  \end{subfigure}
    \hfill
  \begin{subfigure}[b]{0.47\textwidth}
    \includegraphics[width=\textwidth]{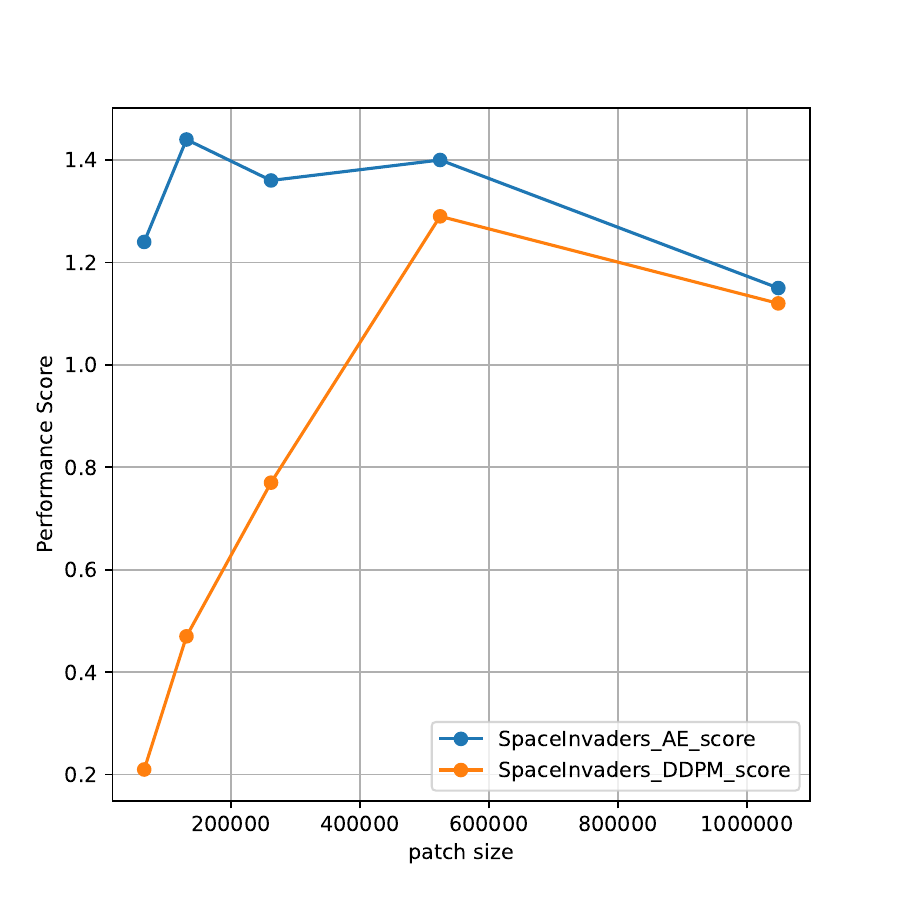}
    \caption{SpaceInvaders.}
    \label{fig:spaceinvaders}
  \end{subfigure}
  \caption{Impact of different autoencoder patch sizes on generation performance in large parameter settings. Both the autoencoder and the diffusion model were trained for 20,000 and 80,000 iterations, respectively.}
  \label{fig:combined}
\end{figure}

In this section, we continue discussing the effect of different autoencoder patch sizes on generation performance in large parameter settings. As shown in Figure~\ref{fig:combined}, we illustrate the AE performance score and DDPM performance score in the Assault and SpaceInvaders environments. Both models were trained with 20,000 iterations for the autoencoder and 80,000 iterations for the diffusion model. The AE performance score represents the ratio of the returns of parameters reconstructed by the autoencoder to the returns of the original parameters, while the DDPM performance score represents the ratio of the returns of parameters generated by the diffusion model to the returns of the original parameters.
In large parameter settings, we experimented with larger patch sizes. Surprisingly, we found that our autoencoder could maintain a high compression rate when transforming the input parameter patch sequence into latent representations, compressing over 1,000,000 dimensions to 32 dimensions, while still achieving an AE performance score greater than 1. Additionally, we observed that as the autoencoder patch size increased, the DDPM performance score also improved. This indicates that a higher compression rate of the autoencoder facilitates better learning by the diffusion model. Therefore, we recommend selecting a higher autoencoder patch size, as long as the AE score remains consistently greater than 1, to enhance the learning efficiency of the diffusion model.

\FloatBarrier
\subsection*{Impact of different diffusion model patch sizes}
We also experimented with different patch sizes for the diffusion model to evaluate their impact on generation performance, as shown in Table~\ref{tab_ddpm_patchsize}. Unlike the autoencoder patch size, we found that the patch size of the diffusion model has minimal effect on generation performance. Various settings consistently resulted in satisfactory parameter generation.

\begin{table}[!htbp]
\centering
\caption{Average scores with different diffusion model patch sizes}
\begin{tabular}{@{}lcccccc@{}}
\toprule
Environment & 2 & 4 & 8 & 16 & 32 \\
\midrule
MountainCar & $1.01$ & $1.03$ & $0.99$ & $1.02$ & $1.04$ \\
Ant & $1.33$ & $1.27$ & $1.5$ & $1.5$ & $1.27$ \\
Assault & $1.07$ & $0.99$ & $1.06$ & $1.17$ & $1.08$ \\
\bottomrule
\end{tabular}
\label{tab_ddpm_patchsize}
\end{table}

\FloatBarrier
\subsection*{Training Recipe}
Table~\ref{tab_hyperparameter} presents the hyperparameters of DPPG in both small parameter and large parameter settings.
We recommend adjusting the autoencoder patch size and the training iterations of the diffusion model according to the specific requirements of different tasks.

\begin{table}[!htbp]
\centering
\caption{Training settings comparison for small and large parameters.}
\small
\setlength{\tabcolsep}{4pt}
\begin{tabularx}{\linewidth}{@{}>{\raggedright\arraybackslash}Xll@{}}
\toprule
\textbf{Training Setting} & \textbf{Small Parameters} & \textbf{Large Parameters} \\
\midrule
$K$ & 200 & 200 \\
batch size & 200 & 200 \\
\midrule
\multicolumn{3}{l}{\textbf{Autoencoder}} \\
\midrule
optimizer & AdamW & AdamW \\
learning rate & 3e-4 & 3e-4 \\
training iterations & 20,000 & 20,000 \\
optimizer momentum & betas=(0.9, 0.999) & betas=(0.9, 0.999) \\
weight decay & 2e-6 & 2e-6 \\
$\xi_v$, noise added on the input parameters & 0.001 & 0.001 \\
$\xi_z$, noise added on the latent representations & 0.1 & 0.1 \\
patch size & 2048 & 524288 \\
$d_{\text{model}}$ & 32 & 32 \\
depth (Transformer blocks) & 3 & 3 \\
dim feedforward & 256 & 256 \\
\midrule
\multicolumn{3}{l}{\textbf{Diffusion Model}} \\
\midrule
optimizer & AdamW & AdamW \\
learning rate & 3e-4 & 3e-4 \\
training iterations & 40,000 & 80,000 \\
optimizer momentum & betas=(0.9, 0.999) & betas=(0.9, 0.999) \\
weight decay & 2e-6 & 2e-6 \\
EMA $\beta$ & 0.9999 & 0.9999 \\
betas start & 1e-4 & 1e-4 \\
betas end & 2e-2 & 2e-2 \\
betas schedule & linear & linear \\
$T$, maximum time steps in training & 1000 & 1000 \\
patch size & 16 & 16 \\
depth (Transformer blocks) & 2 & 2 \\
hidden size & 32 & 32 \\
\bottomrule
\end{tabularx}
\label{tab_hyperparameter}
\end{table}

\FloatBarrier
\end{document}